%% file: main_ieee.tex
\documentclass[letterpaper, 10 pt, conference]{IEEEtran}
\pdfoutput=1    

\usepackage{xcolor}
\definecolor{wong-black}        {HTML}{000000}
\definecolor{wong-lightorange}  {HTML}{E69F00}
\definecolor{wong-lightblue}    {HTML}{56B4E9}
\definecolor{wong-green}        {HTML}{009E73}
\definecolor{wong-yellow}       {HTML}{F0E442}
\definecolor{wong-darkblue}     {HTML}{0072B2}
\definecolor{wong-darkorange}   {HTML}{D55E00}
\definecolor{wong-pink}         {HTML}{CC79A7}

\usepackage[accsupp]{axessibility}  
\usepackage[margin=0.75in]{geometry} 
\usepackage{afterpage}
\usepackage{todonotes}

\usepackage{url}

\usepackage{hyperref} 

\hypersetup{
    colorlinks=true,
    citecolor=wong-green,
    linkcolor=wong-darkblue,
    filecolor=wong-pink,      
    urlcolor=wong-black,
    pdfpagemode=FullScreen,
    }

\newcommand{\secref}[1]{Sec.~\ref{#1}}

\usepackage{cite}
\usepackage{comment}
\usepackage{amsmath,amssymb,amsfonts}
\usepackage{adjustbox}
\usepackage{algorithmic}
\usepackage{graphicx}
\usepackage{booktabs}
\usepackage{textcomp}
\usepackage{hyperref}
\usepackage{dpfloat}
\usepackage{float}
\usepackage{multirow}
\restylefloat{table}
\usepackage{graphicx}
\usepackage{threeparttable}

\def\BibTeX{{\rm B\kern-.05em{\sc i\kern-.025em b}\kern-.08em
    T\kern-.1667em\lower.7ex\hbox{E}\kern-.125emX}}
    
\newcommand\nnfootnote[1]{  
  \begin{NoHyper}
  \renewcommand\thefootnote{}\footnote{#1}%
  \addtocounter{footnote}{-1}%
  \end{NoHyper}
}

\begin{document}


\title{A Survey on Intermediate Fusion Methods for Collaborative Perception Categorized by Real World Challenges}


\author{\IEEEauthorblockN{Melih Yazgan\IEEEauthorrefmark{2}\IEEEauthorrefmark{3}\textsuperscript{\textasteriskcentered},
Thomas Graf\IEEEauthorrefmark{3}\textsuperscript{\textasteriskcentered},
Min Liu\IEEEauthorrefmark{3}\textsuperscript{\textasteriskcentered}, Tobias Fleck\IEEEauthorrefmark{2}\IEEEauthorrefmark{3} and J. Marius Zöllner\IEEEauthorrefmark{2}\IEEEauthorrefmark{3}}

\IEEEauthorblockA{\IEEEauthorrefmark{2}FZI Research Center for Information Technology, Germany\\
\IEEEauthorblockA{\IEEEauthorrefmark{3}Karlsruhe Institute of Technology, Germany\\}} }

\maketitle

\nnfootnote{\textasteriskcentered~These authors contributed equally.} \\



\begin{abstract}
\input{sections/0_abstract}
\end{abstract}




\input{sections/1_introduction}

\input{sections/3_challenges}

\input{sections/4_conclusion}
\input{sections/5_acknowledgment}

{\small
\bibliographystyle{IEEEtran}
\bibliography{references}
}

\end{document}

%% file: sections/0_abstract.tex
This survey analyzes intermediate fusion methods in collaborative perception for autonomous driving, categorized by real-world challenges. We examine various methods, detailing their features and the evaluation metrics they employ. The focus is on addressing challenges like transmission efficiency, localization errors, communication disruptions, and heterogeneity. Moreover, we explore strategies to counter adversarial attacks and defenses, as well as approaches to adapt to domain shifts. The objective is to present an overview of how intermediate fusion methods effectively meet these diverse challenges, highlighting their role in advancing the field of collaborative perception in autonomous driving.


%% file: sections/1_introduction.tex
\section{Introduction}
\label{sec:introduction}

Highly automated vehicles greatly benefit from their many cameras and 3D sensors. While individual perception integrates outputs from various sensors on the ego vehicle for independent object detection\cite{liang2018deep}, it faces constraints in complex scenarios characterized by occlusions, distant objects, and sensor noise\cite{li_nlos_2023}. Collaborative perception can mitigate these limitations by gathering additional perceptual information from other vehicles and infrastructure, a process facilitated by Vehicle-to-Everything (V2X) communication methods\cite{zhou_evolutionary_2020}. This includes Vehicle-to-Vehicle (V2V)\cite{wang_v2vnet_2020} and interactions with roadside infrastructures, referred to as Vehicle-to-Infrastructure (V2I)\cite{bai_pillargrid_2022}, both of which contribute to the optimization of safety, reliability, and efficiency in autonomous driving\cite{hobert_enhancements_2015}.

To implement collaborative perception, it is crucial to integrate data from collaborators into the ego vehicle through fusion techniques. There are three main fusion types: Early Fusion, Late Fusion, and Intermediate Fusion.
Early fusion involves sharing raw sensing data at the input level into a unified coordinate system\cite{chen_cooper_2019}. Although this approach ensures a high level of accuracy, the volume of data to be transmitted is excessively large and sensitive to noise and delay \cite{xu_v2x-vit_2022}. Realizing robust raw data fusion by using current V2X communication capabilities is impossible.
Late fusion methods transmit detection results, representing the output of deep learning models \cite{rauch_car2x-based_2012}. This approach minimizes communication overhead due to the reduced data size, but the associated gains in detection accuracy are constrained.
In intermediate fusion method, feature maps generated by deep learning models are shared \cite{chen_f-cooper_2019}. This feature-level sharing approach enables achieving a remarkably high level of accuracy, accompanied by a reduction in communication cost \cite{wang_v2vnet_2020}. Moreover, the shared intermediate features exhibit increased resilience to GPS inaccuracies and communication latencies\cite{xu_v2x-vit_2022,lei_latency-aware_2022}.

Previous research \cite{han_collaborative_2023, liu_towards_2023, huang_v2x_2023} have provided overviews of various methods and datasets related to collaborative perception. However, these works cover a broad spectrum of topics and deal with individual concepts only on the surface.

This work focuses on intermediate-level collaborative perception methods and provides a categorized overview of these methods into real-world challenges.

The paper is organized as follows: Section \ref{sec:challenges_gaps} systematically analyzes the challenges, and presenting state-of-the-art collaboration methods in Tables \ref{tab:transmission_efficiency}-\ref{tab:perception_domain_shift}. Section \ref{sec:conclusion} concludes with a summary and discussion about open challenges for future research.

%% file: sections/3_challenges.tex
\section{Challenges and Gaps}
\label{sec:challenges_gaps}

\subsection{Transmission Efficiency}
\label{sec:transmission_efficiency}
\begin{table*}
    \centering
    \begin{threeparttable}
    \caption{Overview of the methods for transmission efficiency, clustered by strategy}
    \label{tab:transmission_efficiency}
    \small 
    \begin{tabular}{p{1.2cm}p{2.7cm}p{1cm}p{1cm}p{7.5cm}p{0.8cm}}
    \toprule
    \textbf{Category} & \textbf{Method} & \textbf{Year} & \textbf{Sensors} & \textbf{Key Features} & \textbf{Code} \\ 
    \midrule
    C & V2VNet\cite{wang_v2vnet_2020} & 2020 & L & Adapted variational image compression & - \\
    & DiscoGraph\cite{li_learning_2022} & 2022 & L & 1x1 convolutional autoencoder & \href{https://github.com/ai4ce/DiscoNet}{Link} \\ 
    & AttFusion\cite{xu_opv2v_2022} & 2022 & L & Encoder with multiple 2D convolutions & - \\
    & V2X-ViT\cite{xu_v2x-vit_2022} & 2022 & L & Multiple $1\times1$ convolutions & \href{https://github.com/DerrickXuNu/v2x-vit}{Link} \\
    & FFNet\cite{yu_vehicle-infrastructure_2023} & 2023 & L & Compression combined with feature flow prediction & \href{https://github.com/haibao-yu/FFNet-VIC3D}{Link} \\
    & VIMI\cite{wang_vimi_2023} & 2023 & C & Two-stage compression approach & - \\
    & ScalCompress\cite{yuan_scalable_2023} & 2023 & C & Scalable compression model & - \\
    \midrule
    AS & Who2com \cite{liu_who2com_2020} & 2020 & C & Learnable 1D handshake mechanism & - \\
    & When2com\cite{liu_when2com_2020}  & 2020 & C & Learnable communication graph & - \\
    \addlinespace
    AS + FS & Where2comm\cite{hu_where2comm_2022} & 2022 & L,C & Communication graph by using spatial confidence map & \href{https://github.com/MediaBrain-SJTU/where2comm}{Link} \\
    \addlinespace
    FS & Cooperverse \cite{bai_cooperverse_2023} & 2023 & L & Dynamic feature sharing & - \\
    & CoCa3D\cite{hu_collaboration_2023} & 2023 & C & Critical information selection & - \\
    & Learn2com\cite{wang_collaborative_2023} & 2023 & L & Attention-based feature weighting & - \\
    & How2comm \cite{yang_how2comm_2023}& 2023 & L & Choosing most relevant information& \href{https://github.com/ydk122024/How2comm}{Link} \\
    & CenterCoop\cite{zhou_centercoop_2023} & 2023 & L & Preventing duplicate information in feature map & - \\
    & BM2CP\cite{zhao_bm2cp_2023}  & 2023 & L,C & Critical feature selection through confidence map & \href{https://github.com/byzhaoAI/BM2CP}{Link} \\
    & UMC\cite{wang_umc_2023} & 2023 & L & Entropy-based information selection & - \\
    \midrule
    AS + C & MASH\cite{glaser_overcoming_2021} & 2021 & C & Learnable 2D handshake mechanism, 1x1 conv. encoders  & - \\
    \addlinespace
    FS + C & FPV-RCNN\cite{yuan_keypoints-based_2022}  & 2022 & L & Feature points proposals, 3D data compression & \href{https://github.com/YuanYunshuang/FPV_RCNN}{Link} \\
    & Slim-FCP\cite{guo_slim-fcp_2022} & 2022 & L & Removing irrelevant features, encoder architecture & - \\
    & What2comm \cite{yang_what2comm_2023}& 2023 & L & Importance maps, 1x1 convolutional encoder & - \\
    \bottomrule
    \end{tabular}
    \begin{tablenotes}[flushleft]
        \footnotesize
        \item[1] Categories: Compression (C), Agent Selection (AS), Feature Selection (FS)
        \item[2] Input: Camera (C), Lidar (L)
    \end{tablenotes}
    \end{threeparttable}
\end{table*}

Addressing transmission issues due to bandwidth and computing limitations in collaborative perception is crucial. Recent advancements, presented in Table \ref{tab:transmission_efficiency}, focus on reducing communication bandwidth and transmission latency while maintaining detection performance, achieved through compression methods, selective communication, or a combination of both. Collaborative perception generally requires trade-offs between detection quality and the needed transmission bandwidth.

\textbf{Compression Methods} try to condense the raw input data into an intermediate representation that still encodes the necessary information to perform a downstream task, e.g., object detection. Compression models are typically evaluated by investigating two metrics in parallel: a downstream task metric like the object detection quality and the data transmission size. While data transmission size is sometimes represented on an absolute(logarithmic) scale for visualization purposes\cite{wang_v2vnet_2020,xu_opv2v_2022}, the actual metrics of importance are the Average Byte (AB) and the compression ratio. The AB metric, calculated by representing each element in the transmitted data as a 32-bit float, offers a precise measure of the transmission cost, excluding ancillary data like calibration files and timestamps\cite{yu_vehicle-infrastructure_2023}. Alongside this is the compression ratio, which is the ratio between the input data size and the output size of the compression model.
V2VNet \cite{wang_v2vnet_2020} employs a compression approach where sensor data is first transformed into a condensed intermediate form. This data is then compressed using a Convolutional Neural Network (CNN) and adapted variational image compression algorithm. The process concludes with quantization and entropy encoding. Li et al.\cite{li_learning_2022} presents a student-teacher approach named DiscoGraph. A notable aspect of DiscoGraph is its operational efficiency during inference, requiring only the student model, DiscoNet. It reduces communication volume by employing $1\times1$ convolutional autoencoder and outperforms V2VNet\cite{wang_v2vnet_2020} by higher compression rates in detection accuracy. AttFusion \cite{xu_opv2v_2022} maintains performance at high compression rates. An encoder architecture, featuring multiple 2D convolutions and max pooling, is applied to compress information into a feature map. Experimental results indicate that at a compression rate of 4096, the performance decreases by approximately 3\%, which remains superior to early and late fusion methods. V2X-ViT \cite{xu_v2x-vit_2022} utilize multiple $1\times1$ convolutions for efficient feature map compression, demonstrating high detection performance at various compression rates against \cite{wang_v2vnet_2020,li_learning_2022,xu_opv2v_2022}.
FFNet\cite{yu_vehicle-infrastructure_2023} incorporates a feature flow prediction module with compression. Two compressors are employed for the feature and its derivative, each utilizing three Conv-Bn-ReLU blocks. It reduces the transmission cost by maintaining better detection accuracy than early, late fusion, and intermediate methods such as \cite{wang_v2vnet_2020,li_learning_2022}.
VIMI \cite{wang_vimi_2023} framework utilizes a two-stage compression approach. It employs a compression module incorporating both channel and spatial compression blocks. The transmission cost decreases to a level comparable to late fusion. Scalable Feature Compression \cite{yuan_scalable_2023} further enhances transmission efficiency by employing scalable multiple channel reduction modules. It compresses intermediate features into a layered bitstream. These modules include a $1\times1$ convolution layer for channel reduction before entropy coding and a $5\times5$ convolution layer for channel and spatial dimension reduction of the lower scalable layers.

\textbf{Selective Strategies} try to optimize the bandwidth and channel usage by carefully selecting agents to communicate with and what type of information to send to other agents. The evaluation metrics except compression ratio are the same as we described in compression. 
Who2com \cite{liu_who2com_2020} introduces a three-stage handshake communication mechanism that divides the communication process into request, match, and connect stages. While Who2com operates on the premise that every agent must consistently communicate with one other agent, which leads to inefficient bandwidth usage, When2com\cite{liu_when2com_2020} presents a model that learns to choose when to communicate and to construct communication groups. In the case of a single requesting agent, When2com outperforms Who2com by achieving a 50\% reduction in bandwidth requirements and a higher detection accuracy. Where2comm \cite{hu_where2comm_2022} minimizes bandwidth using a spatial confidence-aware system. The system utilizes the spatial confidence map to identify the most informative spatial areas in the feature map and determines the optimal collaboration partners. Additionally, it only communicates non-zero features along with their corresponding indices. It shows better detection accuracy-bandwidth performance compared to methods\cite{wang_v2vnet_2020,li_learning_2022,xu_v2x-vit_2022,liu_when2com_2020}.
Cooperverse \cite{bai_cooperverse_2023} employs dynamic feature sharing to optimize communication bandwidth by adjusting shared feature cells based on specific bandwidth requirements. Additionally, it filters to prioritize features based on their spatial distance from the sensor. CoCa3D \cite{hu_collaboration_2023} optimizes communication efficiency by selecting the optimal messages and demonstrates superior performance against methods \cite{hu_where2comm_2022,xu_v2x-vit_2022,li_learning_2022,wang_v2vnet_2020}. Learn2com \cite{wang_collaborative_2023} learns to represent feature importance through attention weights, utilizing these weights during inference to determine the most suitable infrastructure for communication. A further framework that reduces communication volume by choosing the messages that contain the most relevant information is How2comm\cite{yang_how2comm_2023}. This method introduces a mutual information-aware communication framework that implements spatial-channel attention queries. A spatial query examines the features, while a channel query selects the most valuable channels. It then identifies the required messages and integrates them into a feature map. It shows better detection accuracy-bandwidth performance compared to methods \cite{wang_v2vnet_2020,xu_opv2v_2022,li_learning_2022,xu_v2x-vit_2022}.
Differently, CenterCoop \cite{zhou_centercoop_2023} addresses overlapping sensor observations by encoding Bird's Eye View (BEV) frames into compact representations. The method prevents duplicate detections arising from overlapping sensor features. Compared to Where2comm\cite{hu_where2comm_2022}, it shows slightly better detection performance in drastically reduced bandwidth. Another multi-model approach, BM2CP \cite{zhao_bm2cp_2023}, achieves communication efficiency by selectively transmitting only the most critical features identified through a binary confidence mask. It shows better performance than the methods \cite{wang_v2vnet_2020,xu_v2x-vit_2022,hu_where2comm_2022}.
UMC \cite{wang_umc_2023} reduces communication costs by employing an entropy-based selection mechanism that transmits only the most informative regions from multi-resolution data. This approach ensures efficient bandwidth usage by filtering out lower-quality regions, effectively balancing the transmission of broad-scale and detailed information while maintaining superior detection accuracy-bandwidth performance compared to other methods.

\textbf{Combined Approaches} aim to enhance both efficiency and
performance by integrating various methods. In our review, we explore the integration of compression and selection. MASH\cite{glaser_overcoming_2021} combine agent selection and compression, using a 2D spatial handshaking mechanism based on Who2com\cite{liu_who2com_2020} for information exchange and bandwidth management, while 1x1 convolutional encoders minimize communication volume. FPV-RCNN \cite{yuan_keypoints-based_2022} combines feature selection from generated proposals with 3D data compression, transmitting less data than a 2D BEV fusion while maintaining superior detection performance. By employing a channel-wise encoder and decoder to reduce feature maps from 128 to 32 channels, Slim-FCP \cite{guo_slim-fcp_2022} effectively filters out non-essential features. It combines attention-weight and semantic-information channel selection to minimize redundancy and an irrelevant feature remover to eliminate non-critical features. What2comm \cite{yang_what2comm_2023} framework, a feature decoupling mechanism is employed to separate data into exclusive and common features. Following this, importance maps identify the most critical regions within these features. After this identification process, a compression module is integrated to compress these selected regions efficiently. Compared to model Where2comm\cite{hu_where2comm_2022}, it attains a superior performance-bandwidth trade-off across all tested bandwidth options.


\subsection{Localization and Pose Error}
\label{sec:loc_pose_error} 
\begin{table*}
    \centering
    \begin{threeparttable}
    \caption{Overview of the methods for localization and pose error, clustered by strategy}
    \label{tab:localization_pose_errors}
    \small 
    \begin{tabular}{p{1.2cm}p{2.8cm}p{0.8cm}p{1cm}p{8.8cm}p{0.8cm}}
    \toprule
    \textbf{Category} & \textbf{Method} & \textbf{Year} & \textbf{Sensors} & \textbf{Key Features} & \textbf{Code} \\ \midrule
    EC & RobustV2VNet\cite{vadivelu_learning_2020} & 2020 & L & Outlier pose error correction & - \\
     & FPV-RCNN\cite{yuan_keypoints-based_2022} & 2022 & L & Maximum consensus algorithm for precise error estimation & \href{https://github.com/YuanYunshuang/FPV_RCNN}{Link} \\
     & CoAlign\cite{lu_robust_2023} & 2022 & L & Aligns relative poses among agents & \href{https://github.com/yifanlu0227/CoAlign}{Link} \\
     & FeaCo\cite{Gu2023FeaCoRR} & 2023 & L & Refining pose-error through spatial mapping and precision alignment & \href{ https://github.com/jmgu0212/FeaCo.git}{Link}\\
     & BM2CP\cite{zhao_bm2cp_2023} & 2023 & L,C & Pose adjustment before feature communication & \href{ https://github.com/byzhaoAI/BM2CP}{Link}\\
    \midrule
    STC & V2X-ViT\cite{xu_v2x-vit_2022} & 2022 & L & Multi-scale window attention and spatio-temporal correction  & \href{https://github.com/DerrickXuNu/v2x-vit}{Link}\\
     & Select2Col\cite{liu_select2col_2023}  & 2023 & L & Using historical prior hybrid attention & \href{https://github.com/huangqzj/Select2Col/}{Link} \\
     & SCOPE\cite{yang_spatio-temporal_2023} & 2023 & L & Learning spatio-temporal correlations in historical contexts & \href{https://github.com/starfdu1418/SCOPE}{Link} \\
     & What2comm\cite{yang_what2comm_2023}& 2023 & L & Integration of exclusive representations and historical ego cues & - \\
     & How2comm\cite{yang_how2comm_2023} & 2023 & L & Capturing perceptually critical and temporally valuable information & \href{https://github.com/ydk122024/How2comm}{Link} \\
    \bottomrule
    \end{tabular}
    \begin{tablenotes}[flushleft]
        \footnotesize
        \item[1] Category: Error Correction (EC), Spatio-Temporal Collaboration (STC)
        \item[2] Sensors: Camera (C), Lidar (L)
    \end{tablenotes}
    \end{threeparttable}
\end{table*}

Localization and pose errors in collaborative perception can lead to feature map misalignment and performance drops. These effects result from errors of the used localization systems, e.g., GPS inaccuracies, sensor noise, temporal misalignment, synchronization errors, or pose drift. Methods addressing these issues are categorized in Table \ref{tab:localization_pose_errors}. All these models are typically evaluated using a quality metric of a downstream task, e.g., object detection accuracy, by comparing it to the detection quality under noise. Noisy Settings are generally generated from a Gaussian distribution. Only in RobustV2Vnet\cite{vadivelu_learning_2020} is rotational noise generated with von Mises distribution.

\textbf{Error Correction} is a categorization focused on methods that proactively address and rectify errors at each agent or sensor level before the information fusion. Vadivelu et al.\cite{vadivelu_learning_2020} proposed RobustV2VNet, an end-to-end learnable pose error correction network. This framework features a pose regression module designed to correct pose errors by learning a parameter for adjusting noisy relative transformations. Additionally, they introduce a consistency module that employs a Markov random field to refine these predictions, ensuring a globally consistent absolute pose across all agents. FPV-RCNN\cite{yuan_keypoints-based_2022} introduced a localization error correction mechanism
that classifies keypoints into various environmental elements and selectively shares their coordinates. The crucial part of the module is the maximum consensus algorithm, used to identify corresponding vehicle centers and pole points, aiding in precise localization error estimation. In contrast, CoALign\cite{lu_robust_2023} introduced an agent-object pose graph optimization without requiring accurate pose supervision in training. It aligns relative poses among agents and detected objects, promoting pose consistency. It operates without requiring accurate pose supervision in training, showcasing strong generalization to various levels of pose errors.
FeaCo\cite{Gu2023FeaCoRR} presents a pose-error rectification module to combat pose errors in noisy settings. The module operates in two pivotal stages. The first, the localization stage, involves creating a spatial confidence map to facilitate the selection of critical regions for proposal matching on the feature map, a crucial step for alignment precision. In the second stage, the matching phase, it calculates a fine-grained transformation matrix, which is essential for aligning feature maps and ensuring consistent synchronization among agents. Most recently, BM2CP\cite{zhao_bm2cp_2023} introduced BM2CP-robust, which adjusts the relative pose prior to sensor fusion. It calculates local bounding boxes and their uncertainty for each agent based on LiDAR voxel features. Following this, it utilizes the agent-object pose graph optimization from CoAlign\cite{lu_robust_2023} for each agent to enhance the refinement of the relative pose.

\textbf{Spatio Temporal Collaboration} plays a crucial role in error correction and addressing misalignments, particularly at the information fusion stage. This category of methodologies is designed to tackle the challenges of integrating data from multiple sources, ensuring that spatial and temporal discrepancies are effectively corrected. V2X-ViT\cite{xu_v2x-vit_2022} addresses GPS-related localization errors using its Multi-Scale window attention module (MSwin), which captures diverse visual information for spatial accuracy. Additionally, its Spatio-Temporal Correction Module (STCM) corrects global misalignments by warping feature maps for better data alignment. Differently, Select2Col\cite{liu_select2col_2023} introduced a Historical Prior Hybrid Attention (HPHA) information fusion algorithm. It utilizes a short-term attention module to capture the correlation of semantic information from the temporal dimension. HPHA determines incorrect features using historical feature information, reducing spatial localization noise.
Another method that uses historical information is SCOPE\cite{yang_spatio-temporal_2023}. It ensures that the ego agent integrates confidence-aware spatial information by utilizing a cross-attention module. Concurrently, an adapted LSTM module extracts contextual dependencies from historical frames. Continuing this trend, What2comm\cite{yang_what2comm_2023} introduced a Spatial Attention Integration (SAI) and a Temporal Context Aggregation (TCA) module. The SAI component, leveraging a cross-attention module, effectively aligns position embeddings of features with their associated importance maps. This process allows for the extraction of perceptually crucial information. Concurrently, the TCA module integrates historical feature representations via a gating mechanism, substantially mitigating collaboration noise. Its robustness presented compared to V2X-ViT\cite{xu_v2x-vit_2022}, V2VNet\cite{wang_v2vnet_2020}, and Where2Comm\cite{hu_where2comm_2022}. Unlike previous methods that relied heavily on historical data, How2comm\cite{yang_how2comm_2023} presented a Spatio-Temporal Collaboration Transformer (STCFormer) along with a flow-guided delay compensation strategy. The STCFormer integrates spatial and temporal data from multiple agents, enhancing perception accuracy even amidst spatial misalignments caused by localization errors. The flow-guided strategy predicts future states of collaborating agents to maintain feature alignment.

\subsection{Communication Issues}
\label{sec:comm_issues}
    \begin{table*}
    \centering
    \begin{threeparttable}
    \caption{Overview of the methods for compensating communication issues, clustered by challenge}
    \label{tab:communication_issues}
    \small 
    \begin{tabular}{p{1.2cm}p{2.9cm}p{1.0cm}p{1cm}p{7.5cm}p{0.8cm}}
    \toprule
    \textbf{Category} & \textbf{Method} & \textbf{Year} & \textbf{Sensors} & \textbf{Key Features} & \textbf{Code} \\ 
    \midrule
    L & V2X-ViT\cite{xu_v2x-vit_2022} & 2022 & L & Delay-aware positional encoding & \href{https://github.com/DerrickXuNu/v2x-vit }{Link} \\
    & SyncNet\cite{lei_latency-aware_2022} & 2022 & L & Capturing spatial features at multiple scales & - \\
    & FFNet\cite{yu_vehicle-infrastructure_2023} & 2023 & L & Prediction of future features using feature flow & \href{https://github.com/haibao-yu/FFNet-VIC3D }{Link} \\
    & CoBEVFlow\cite{wei_asynchrony-robust_2023} & 2023 & L, C & Feature map estimation by applying flow map & \href{https://github.com/MediaBrain-SJTU/CoBEVFlow}{Link}  \\
    \midrule
    LC & V2VAM+LCRN \cite{li_learning_2023} & 2023 & L & Recovering incomplete features & - \\
    & V2X-INCOP\cite{ren_interruption-aware_2023} & 2023 & L & Using historical data to recover lost information & - \\
    \midrule
     NLOS & APMM+MoHeD\cite{li_nlos_2023} & 2023 & L & Identifying blind zones and calculating perception benefits & - \\
    \bottomrule
    \end{tabular}
    \begin{tablenotes}[flushleft]
        \footnotesize
        \item[1] Category: Latency (L), Lossy Communication (LC), NLOS (Non-Line-of-Sight)
        \item[2] Sensors: Lidar (L), Camera (C)
    \end{tablenotes}
    \end{threeparttable}
    \end{table*}

Communication delays and interruptions, such as network congestion and signal loss, are critical challenges in collaborative perception. Various methods, shown in Table \ref{tab:communication_issues}, have been developed to address these issues, focusing on latency, Lossy Communication (LC), and Non-Line-of-Sight (NLOS) conditions.
    
\textbf{Latency} is defined by the time difference between recording the sensor data in one agent and the time after de-serialization is finished at the second agent. Latency is highly influenced by (de-)serialization speed and inference speed of models. Since latency is inevitable, methods that register multiple sources with different latency are of particular interest. The performance of the introduced models is typically measured by a downstream task quality metric, e.g., object detection accuracy. The method is considered valid if the downstream task metric improves when compensating latency.
V2X-ViT \cite{xu_v2x-vit_2022}, besides transmission efficiency and localization errors, integrates adaptive delay-aware positional encoding, capturing temporal information to ensure accurate synchronization despite delays. Differently, SyncNet\cite{lei_latency-aware_2022} introduces a novel latency-aware collaborative perception system that allows it to serve as a plugin latency compensation module for various collaborative perception methods. It synchronizes perceptual features across agents to a standard time stamp, which enhances collaboration. SyncNet simultaneously estimates real-time features and collaboration attention. This marks a distinctive departure from traditional methods. Historical information is harnessed to achieve synchronization, with dual-branch pyramid Long Short-Term Memory (LSTM) networks capturing spatial features at multiple scales. In addition to historical information, flow maps are utilized to address latency. FFNet\cite{yu_vehicle-infrastructure_2023} proposes a self-supervised approach to train a feature flow generator. The generator predicts future features and compares them with the ego vehicle's sensor data. It employs a linear operation for prediction, effectively addressing temporal fusion errors across different latencies and compensating for uncertainties in latency. Experiments demonstrate the robustness of FFNet to various latencies. To correct the feature errors caused by latency, CoBEVFlow\cite{wei_asynchrony-robust_2023} generates a flow map from BEV. In this process, related regions of interest are matched based on an irregularly sampled sequence of messages sent by the same agent. Using the BEV flow map, the estimated feature map is then determined to reallocate the asynchronous features. Compared to SyncNet\cite{lei_latency-aware_2022}, V2X-ViT\cite{xu_v2x-vit_2022}, CoBEVFlow achieves the best performance under various time delays.
    
\textbf{LC} refers to the data transmission resulting in lost or altered information, which is another significant challenge. LC can lead to losing a data frame from a specific timestamp or just blocks of the data, e.g., spatially correlated subblocks of a feature map or grid structure.
V2VAM+LCRN \cite{li_learning_2023} introduces an LC-aware repair network and a V2V attention module, focusing on recovering incomplete shared features and enhancing vehicle interaction. The effectiveness of the proposed method is shown compared to V2X-ViT\cite{xu_v2x-vit_2022} using object detection accuracy in lossy conditions. V2X-INCOP \cite{ren_interruption-aware_2023} uses historical data to recover information lost during communication interruptions, stabilizing the model's training under imperfect communication conditions. The 3D detection performance of the model in various packet drop rates outperforms the methods such as V2X-ViT \cite{xu_v2x-vit_2022} and Where2Comm \cite{hu_where2comm_2022}.

\textbf{NLOS} conditions, are characterized by the absence of direct visual contact between the transmitter and receiver, are addressed by APMM+MoHeD \cite{li_nlos_2023}. It combines matrix matching, which transforms and broadcasts sensor data into a matrix with vehicle coordinates, enabling the detection of overlapping blind zones and an optimized relay node selection strategy. This strategy prioritizes nodes based on mobility and height, enhancing packet reception, reducing collision risks, and minimizing the need for relay re-selection.

\subsection{Heterogeneity}
\label{sec:model_task_data_Heterogeneity}
\begin{table*}
    \centering
    \begin{threeparttable}
    \caption{Overview of the methods for heterogeneity, clustered by challenge}
    \label{tab:heterogeneity}
    \small 
    \begin{tabular}{p{1.2cm}p{2.2cm}p{1cm}p{1cm}p{7.5cm}p{0.8cm}}
    \toprule
    \textbf{Category} & \textbf{Method} & \textbf{Year} &\textbf{Sensors} & \textbf{Key Features} & \textbf{Code} \\ \midrule
    MH & MPDA\cite{xu_bridging_2023} & 2023 & L & Aligns neural networks  & - \\
    \midrule
    TH & STAR\cite{li_multi-robot_2023} & 2023 & L &  Enabling unified task handling through shared data & \href{https://github.com/coperception/star}{Link} \\ 
     & CORE\cite{wang_core_2023} & 2023 & L & Comprehensive environmental understanding & \href{https://github.com/zllxot/CORE}{Link} \\ \midrule
     
    SH & HPL-ViT\cite{liu_hpl-vit_2023} & 2023 & L &  Feature fusion for diverse LiDAR systems & - \\ 
     & HM-ViT\cite{xiang_hm-vit_2023} & 2023 & L,C & Feature fusion with varying sensors & - \\ 
     & BM2CP\cite{zhao_bm2cp_2023} & 2023 & L,C & Robustness in sensor absence & \href{https://github.com/byzhaoAI/BM2CP}{Link} \\
     & DI-V2X\cite{xiang_di-v2x_2023}& 2023 & L & Standardizing diverse LiDAR sensor inputs & \href{https://github.com/Serenos/DI-V2X}{Link}\\
     & FedBEVT\cite{song_fedbevt_2023}& 2023 & C & Addressing data variability in privacy-sensitive scenarios & \href{https://github.com/rruisong/FedBEVT}{Link} \\ 
    \bottomrule
    \end{tabular}
    \begin{tablenotes}[flushleft]
        \footnotesize
        \item[1] Type: Model Heterogeneity (MH), Task Heterogeneity (TH), Sensor Heterogeneity (SH)
        \item[2] Sensors: Camera (C), Lidar (L)
    \end{tablenotes}
    \end{threeparttable}
\end{table*}
Heterogeneity within collaborative perception presents a multifaceted challenge that spans models, tasks, sensors. While enriching the system with a broad spectrum of information and capabilities, this diversity also introduces significant complexities in integrating and interpreting varied sources and types of information. Methods addressing these issues are summarized in Table \ref{tab:heterogeneity}.

\textbf{Model Heterogeneity} necessitates the alignment of intermediate states from different models before their information can be fused due to the inherent differences in the characteristics of these states across models. Such models are evaluated using a downstream task, e.g., object detection accuracy. 
The MPDA\cite{xu_bridging_2023} framework addresses model heterogeneity by introducing a learnable resizer and sparse cross-domain transformer. The resizer dynamically adjusts feature maps from different agents, aligning spatial resolutions and channel dimensions. The transformer then harmonizes these features, ensuring domain-invariant representations through adversarial training.
    
\textbf{Task Heterogeneity} is a particular case of model heterogeneity where the alignment has to compensate for different target objectives of the different model tasks. Evaluation of such models can be done by using the downstream task objectives. STAR\cite{li_multi-robot_2023} presents a task-agnostic strategy using a spatio-temporal autoencoder. It enables the sharing of spatially subsampled and temporally mixed encoded information, facilitating scene reconstruction from shared data. This method provides a unified base for various tasks, overcoming task discrepancies. On the other hand, CORE\cite{wang_core_2023} adopts a collaborative reconstruction approach. It combines a compression module for efficient feature broadcasting, an attention-based collaboration module for integrating messages, and a reconstruction module synthesizing these inputs into a comprehensive environmental understanding.
    
\textbf{Sensor Heterogeneity} occurs when different sensor modalities are to be fused, but also if different sensor intrinsic properties (e.g., beam count and resolution for LiDAR, intrinsic parameters for a camera) of a single modality are of interest. Performing better than single modalities or single sensors on the downstream task is the evaluation method for these models.
The first study, led by Liu et al.\cite{liu_hpl-vit_2023}, introduces the HPL-ViT, designed for V2V scenarios with diverse LiDAR systems. It employs graph-attention transformers to extract and fuse features specific to each agent, initiated by creating BEV feature maps. Another significant contribution comes from the HM-ViT\cite{xiang_hm-vit_2023}, which spatially transforms and updates features from different sensors through local and global attention mechanisms and outputs final predictions through a vision transformer. Complementing this, the BM2CP\cite{zhao_bm2cp_2023} leverages LiDAR-guided feature selection, integrates depth information from LiDAR and camera for enhanced accuracy, and demonstrates adaptability by maintaining performance despite sensor absence. In addressing the challenges of sensor heterogeneity from a different angle, DI-V2X\cite{xiang_di-v2x_2023} focuses on aligning feature representations from various LiDAR sensors, using domain-mixing instance augmentation and domain-adaptive fusion for consistent 3D object detection. FedBEVT\cite{song_fedbevt_2023} introduces a federated learning framework tailored for BEV perception in road traffic systems. It addresses data variability with camera-specific adjustments and adaptive multi-camera masking. This approach is particularly effective when data privacy is crucial and camera setups vary widely among vehicles.

\subsection{Adversarial Attack and Defense}
    In collaborative perception systems, defending against malicious attacks is vital for maintaining data integrity and system trust. Table \ref{tab:attack_defense} summarizes methods addressing these issues.
    Tu et al.\cite{tu_adversarial_2021} showed that adversarial attacks reduce multi-agent system performance. However, their impact lessens with more collaborative non-attacker agents, highlighting the need for enhanced neural network robustness. ROBOSAC\cite{li_among_2023} introduces a consensus-based defense effective against unknown attack models, especially with more collaborative agents. Zhang et al.\cite{zhang_data_2023} focus on vehicular systems, proposing an anomaly detection system to detect data fabrication attacks by identifying inconsistencies. Complementing these, MADE\cite{zhao_malicious_2023} employs a semi-supervised anomaly detection method to identify and remove malicious agents, proving effective against unknown attacks and controlling false positives.
    \label{sec:adv_attack}
\begin{table*}
    \centering
    \begin{threeparttable}
    \caption{Overview of the attack and defense methods}
    \label{tab:attack_defense}
    \small 
    \begin{tabular}{p{1.2cm}p{2.2cm}p{1cm}p{1cm}p{7.5cm}p{0.8cm}}
    \toprule
    \textbf{Category} & \textbf{Method} & \textbf{Year} & \textbf{Sensors} & \textbf{Key Features} & \textbf{Code} \\
    \midrule
    Attack & AdvAttack \cite{tu_adversarial_2021} & 2021 & L& Adversarial impact reduction with increased agents. & - \\
    \midrule
    Defense & ROBOSAC \cite{li_among_2023}  & 2023 & L & Consensus-based defense & \href{https://github.com/coperception/ROBOSAC}{Link} \\
     & CAD\cite{zhang_data_2023} & 2023 & L & Detection of data fabricated attacks & \href{https://github.com/zqzqz/AdvCollaborativePerception}{Link} \\
     & MADE \cite{zhao_malicious_2023} & 2023 & L & Removing malicious agent  & - \\
    \bottomrule
    \end{tabular}
    \begin{tablenotes}[flushleft]
        \footnotesize
        \item[1] Sensors: Lidar (L), Camera (C)
    \end{tablenotes}
    \end{threeparttable}
\end{table*}
    
\subsection{Domain Shift}
\label{sec:doman_shift}

    In addition to the previously delineated challenges, collaborative perception systems encounter the often understated domain shift problem due to the domain gap between simulation-based and real-world data. Table \ref{tab:perception_domain_shift} summarizes the methods addressing these issues. The S2R-ViT\cite{li_s2r-vit_2023} takes a novel approach to this issue, concentrating on deployment and feature gaps. It integrates a vision transformer and a feature adaptation method, both adept at handling uncertainties in data transition from simulations to real-world scenarios. The vision transformer focuses on identifying and managing uncertainties, while the feature adaptation technique aligns and adapts simulated data features for real-world applicability. This integration has proven effective in 3D object detection within simulation-to-reality scenarios, surpassing frameworks such as AttFuse\cite{xu_opv2v_2022}, V2VNet\cite{wang_v2vnet_2020}, and V2X-ViT\cite{xu_v2x-vit_2022}.
    Differently from  S2R-ViT, DUSA\cite{kong_dusa_2023} expands the horizon by addressing sim-to-real adaptation and innovatively tackling domain discrepancies between different sensor setups. It employs an approach that adapts to location-specific characteristics of real-world scenarios for integrating essential features alongside a method that ensures consistency of these features when shared among diverse agents. Its effectiveness is further validated in various collaborative 3D detection methods such as V2X-ViT\cite{xu_v2x-vit_2022}, DiscoNet\cite{li_learning_2022}.

\begin{table*}
        \centering
        \begin{threeparttable}
        \caption{Overview of the methods for domain shift}
        \label{tab:perception_domain_shift}
        \small 
        \begin{tabular}{p{1.2cm}p{2.2cm}p{1cm}p{1cm}p{7.5cm}p{0.8cm}}
        \toprule
        \textbf{Category} & \textbf{Method} & \textbf{Year} & \textbf{Sensors} & \textbf{Key Features} & \textbf{Code} \\
        \midrule
        DS& S2R-ViT\cite{li_s2r-vit_2023} & 2023 & L & Addressing deployment and feature gaps & - \\
        & DUSA\cite{kong_dusa_2023} & 2023 & L & Tackling sim-to-real and inter-agent domain gap & \href{https://github.com/refkxh/DUSA}{Link} \\
        \bottomrule
        \end{tabular}
        \begin{tablenotes}[flushleft]
            \footnotesize
            \item[1] Domain Shift (DS)
            \item[2] Sensors: Lidar (L), Camera (C)
        \end{tablenotes}
        \end{threeparttable}
    \end{table*}

%% file: sections/4_conclusion.tex
\section{Conclusion}
\label{sec:conclusion}
We have categorized various intermediate fusion methods for collaborative perception, highlighting their potential and limitations. A predominant focus observed is on transmission efficiency, often evaluated within simulated environments. However, these simulations rarely capture the complexities of real-world scenarios, raising concerns about the real-time capabilities of sophisticated architectures like those based on transformers. Simulations often fail to model accurately the significant fluctuations in latency caused by real-world environmental and infrastructural factors, such as signal obstructions in dense urban areas. Additionally, as the number of agents in a network increases, so does the data traffic, which can lead to congestion and further latency if the network infrastructure lacks scalability.

This situation underscores the critical need for research focused on enhancing the scalability of networks to accommodate the expected rise in Connected and Autonomous Vehicles (CAVs) on the roads. The anticipated increase in CAVs presents both opportunities and challenges. More CAVs could enhance the quality and quantity of data available for collaborative perception, yet the effective selection and utilization of this data remains underexplored. Future research should focus on developing dynamic collaboration graphs to identify the most informative agents and exploring parallel information fusion processing.

The datasets currently used to evaluate these methods often feature only a small number of collaborative agents, failing to represent the scale of real-world operations adequately. Compounded by challenges such as labeling, privacy concerns, and the significant investments required to gather comprehensive real-world datasets, domain shift remains a formidable barrier\cite{yazgan2024collaborative}. Additionally, the integration of diverse sensor configurations poses another critical issue. Different vehicle manufacturers use various sensor setups, leading to a broad spectrum of data types and formats. Most current methods rely on homogeneous sensor configurations for validation, which fails to address the diversity seen in actual deployment scenarios, as indicated in 
Tables \ref{tab:transmission_efficiency}-\ref{tab:perception_domain_shift}. The complexity of integrating these varied data sources is compounded by the different models used for autonomous navigation. This increases the difficulty of interpreting and integrating shared feature maps. This necessitates more sophisticated data integration methods for effective collaboration across diverse sensor and system architectures.

In conclusion, while remarkable progress has been made in intermediate fusion for collaborative perception, the journey toward fully unleashing its potential is fraught with challenges. Addressing these issues demands a multidisciplinary approach, leveraging expertise from communication technology, data fusion, security, and machine learning. We hope our findings facilitate further research on intermediate-level collaborative perception.

%% file: sections/5_acknowledgment.tex
\section{Acknowledgment}
\label{sec:acknowledgment}
This work is developed within the framework of the "Shuttle2X" project, funded by the Federal Ministry for Economic Affairs and Climate Action (BMWK) and the European Union, under the funding code 19S22001B. The authors are solely responsible for the content of this publication.